%% file: example_paper.tex
\PassOptionsToPackage{usenames,dvipsnames}{xcolor}
\documentclass{article}

\usepackage{microtype}
\usepackage{graphicx}
\usepackage{booktabs} 

\usepackage{hyperref}
\usepackage{url}            
\usepackage{amsfonts}       
\usepackage{nicefrac}       
\usepackage{microtype}      
\usepackage{xcolor}         
\usepackage{bm}
\usepackage{multirow}

\usepackage[ruled,vlined]{algorithm2e}

 \usepackage[accepted]{icml2024}

\usepackage{amsmath}
\usepackage{amssymb}
\usepackage{mathtools}
\usepackage{amsthm}

\usepackage{caption}
\usepackage{subcaption}
\usepackage{tabularx}
\usepackage{enumitem}
\usepackage[misc]{ifsym}
\usepackage{tikz}
\usepackage{comment}

\SetCommentSty{mycommfont}

\SetKwInput{KwInput}{Input}                
\SetKwInput{KwOutput}{Output}              

\input{math_commands.tex}

\usepackage[capitalize,noabbrev]{cleveref}

\theoremstyle{plain}

\theoremstyle{definition}

\theoremstyle{remark}

\usepackage[textsize=tiny]{todonotes}

\icmltitlerunning{TIC-TAC: A Framework for Improved Covariance Estimation in Deep Heteroscedastic Regression}

\begin{document}

\twocolumn[
\icmltitle{TIC-TAC: A Framework for Improved Covariance Estimation \\ in Deep Heteroscedastic Regression}



\icmlsetsymbol{equal}{*}

\begin{icmlauthorlist}
\icmlauthor{Megh Shukla}{epfl}
\icmlauthor{Mathieu Salzmann}{epfl,sdsc}
\icmlauthor{Alexandre Alahi}{epfl}
\end{icmlauthorlist}

\icmlaffiliation{epfl}{École Polytechnique Fédérale de Lausanne (EPFL)\newline}
\icmlaffiliation{sdsc}{Swiss Data Science Center (SDSC)}

\icmlcorrespondingauthor{Megh Shukla}{megh.shukla@epfl.ch}

\icmlkeywords{Deep Heteroscedastic Regression, Negative log-likelihood}

\vskip 0.3in
]



\printAffiliationsAndNotice{}  

\begin{abstract}
Deep heteroscedastic regression involves jointly optimizing the mean and covariance of the predicted distribution using the negative log-likelihood. However, recent works show that this may result in sub-optimal convergence due to the challenges associated with covariance estimation. While the literature addresses this by proposing alternate formulations to mitigate the impact of the predicted covariance, we focus on improving the predicted covariance itself. We study two questions: (1) Does the predicted covariance truly capture the randomness of the predicted mean? (2) In the absence of supervision, how can we quantify the accuracy of covariance estimation? We address (1) with a \textit{Taylor Induced Covariance (TIC)}, which captures the randomness of the predicted mean by incorporating its gradient and curvature through the second order Taylor polynomial. Furthermore, we tackle (2) by introducing a \textit{Task Agnostic Correlations (TAC)} metric, which combines the notion of correlations and absolute error to evaluate the covariance. We evaluate TIC-TAC across multiple experiments spanning synthetic and real-world datasets. Our results show that not only does TIC accurately learn the covariance, it additionally facilitates an improved convergence of the negative log-likelihood. Our code is available at \url{https://github.com/vita-epfl/TIC-TAC}
\end{abstract}

\section{Introduction}


\begin{figure*}
\begin{subfigure}{1.45\columnwidth}
    \includegraphics[width=\columnwidth]{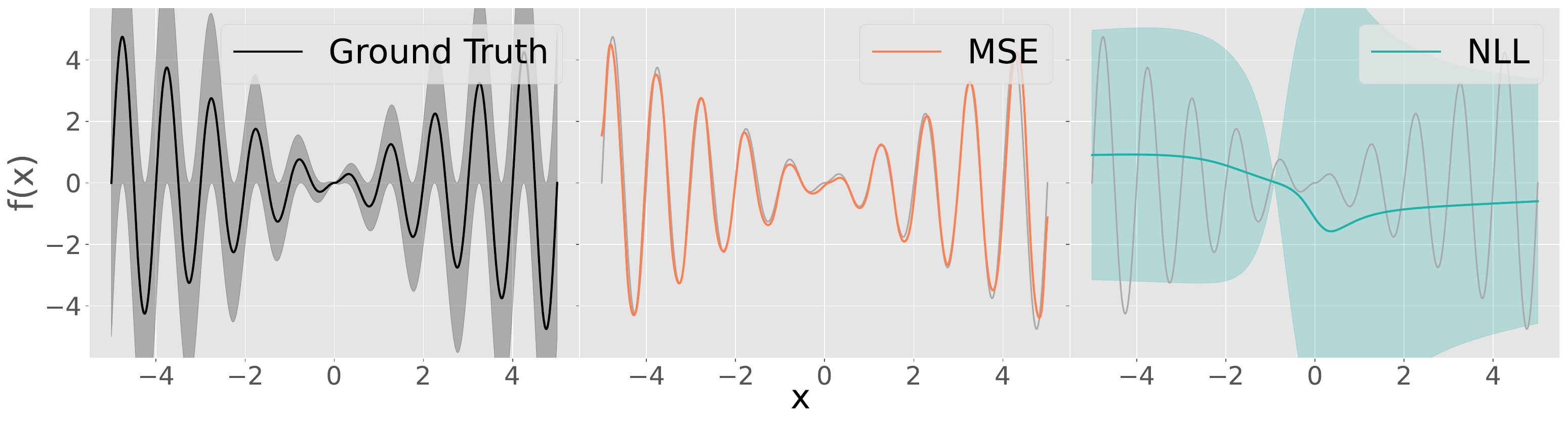}
\end{subfigure}
\hspace{0.15cm}
\begin{subfigure}{0.5\columnwidth}
    \includegraphics[width=\columnwidth]{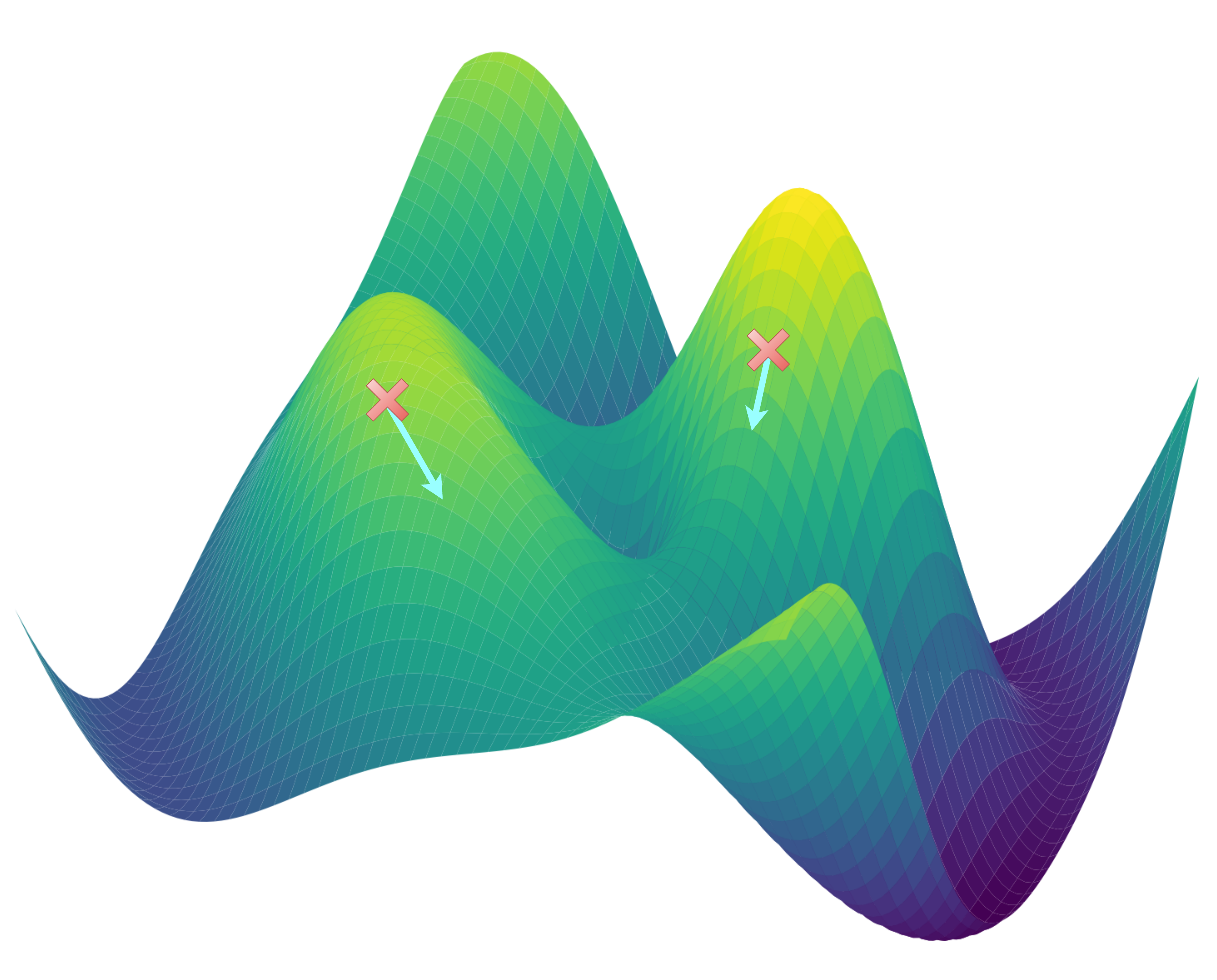}
\end{subfigure}
\caption{\textit{Motivation}. (Left) We learn a varying amplitude sinusoidal with heteroscedastic variance (shaded region). We observe sub-optimal convergence since the predicted variance may be arbitrary and incorrectly minimizes the likelihood.  We address this through a \textit{Taylor Induced Covariance} by tying the randomness of the prediction to its gradient and curvature. (Right) \textit{The gradient and curvature quantify the variation in the prediction within a small neighborhood of the input.}}
\label{fig:main_fig}
\end{figure*}

Modeling the target distribution is an important design choice in heteroscedastic regression. Typically, the target is assumed to follow a multivariate normal distribution, where the true mean and covariance are sample dependent and unknown. Deep heteroscedastic regression learns this distribution by predicting the mean and covariance through two neural networks, which are jointly optimized to minimize the negative log-likelihood. However, recent results in deep heteroscedastic regression show that this joint optimization leads to sub-optimal convergence. 

This challenge is primarily attributed to covariance estimation in heteroscedastic regression \cite{skafte2019reliable}. Recent studies show that the gradient of incorrect variance predictions significantly hinders optimization, and address this by proposing alternate formulations to mitigate its impact during optimization \citep{skafte2019reliable, seitzer2022on, stirn2023faithful, immer2023effective}. While these approaches aim at regularizing the covariance, this begets the question: Can we improve upon the predicted covariance? We argue that the current parameterization for the covariance may not truly explain the randomness of the predicted mean. Indeed, we observe in Figure \ref{fig:main_fig} that in the absence of direct supervision, the predicted variance may take on arbitrary values leading to sub-optimal convergence. Moreover, evaluating the covariance is challenging without ground-truth labels. Optimization metrics such as the likelihood are not a direct measure for the covariance since they also incorporate the performance of the mean estimator.

Hence, this paper studies covariance estimation in deep heteroscedastic regression. We distill the challenges into two problems: (1) How do we model the covariance to explain the randomness of the prediction? (2) How do we evaluate the predicted covariance in the absence of annotations?

\enlargethispage{\baselineskip}

Our first contribution, the \textbf{Taylor Induced Covariance (TIC)}, explains the randomness of the prediction through its gradient and curvature. We develop a closed-form approximation for the covariance of the prediction through its second order Taylor polynomial. Modeling the covariance through the gradient and curvature quantifies the variation in the prediction within a small neighborhood of the input. TIC when learnt through the negative log-likelihood not only captures the underlying correlations but also improves the convergence of the negative log-likelihood.

Our second contribution, the \textbf{Task Agnostic Correlations (TAC)}, addresses the lack of a direct metric to evaluate the covariance. By definition, an accurate covariance correctly estimates the underlying correlations. Hence, given a partial observation of the target, the covariance should accurately update the prediction towards the unobserved target through conditioning of the predicted distribution. Consequently, we quantify TAC as the mean absolute error between the updated prediction and the unobserved target. While the likelihood is a measure of optimization, TAC quantifies the accuracy of the learnt correlations.

We design and perform extensive experiments on synthetic (sinusoidal, multivariate) and real-world (UCI Regression and Human Pose - MPII, LSP) datasets using two metrics: TAC and the likelihood. Our experiments show that TIC outperforms the state-of-the-art baselines in learning correlations across all tasks, demonstrating improved covariance estimation in heteroscedastic regression. Additionally, we also observe that incorporating TIC into the negative log-likelihood improves convergence. Our code and environment are publicly available for reproducibility\footnote{\url{https://github.com/vita-epfl/TIC-TAC}}.

\section{Deep Heteroscedastic Regression}

The goal of heteroscedastic regression is to learn the unknown target distribution $p(Y | X \!=\! \vx)$, which is commonly assumed to be a multivariate normal distribution $\mathcal{N}(\vmu_{Y|X}, \mSigma_{Y|X})$. Typically, deep heteroscedastic regression is performed through minimizing the negative log-likelihood of the predicted distribution $q(\hat{Y} | X \!=\! \vx) = \mathcal{N}(\hat{\vy}, \Cov(\hat{Y} | X))$. This involves the joint optimization of estimators for the mean $\hat{\vy} = f_{\theta}(\vx)$ and the covariance $\Cov(\hat{Y} | X) = g_{\Theta}(\vx)$ over the dataset \cite{nix1994estimating, kendall2017uncertainties}:
\begin{equation}
\resizebox{0.9\columnwidth}{!}{$
    \mathbb{E}_{p(X, Y)} \bigg[ \log \; \Bigl| \Cov(\hat{Y} | X) \Bigr| + (\vy  - \hat{\vy})^T \,\, \Cov(\hat{Y} | X)^{-1} \,\, (\vy  - \hat{\vy}) \bigg].
$}
\label{eq:nll}
\end{equation}
The advantage of deep heteroscedastic regression over its non-parametric counterparts like the Gaussian Process \cite{gp_heteroscedastic} is the ability to extract complex features from inputs such as images. This has lead to its adoption across various paradigms such as active learning \cite{houlsby2011bayesian, gal2017deep}, uncertainty estimation \cite{gal2016dropout, kendall2017uncertainties, lakshminarayanan2017simple, russell2021multivariate}, image reconstruction \cite{dorta2018structured}, human pose estimation \cite{ajain, nakka2023understanding, Tekin_2017_ICCV}, and other vision tasks \citep{lu2022few, simpson2022learning, 8461047, bertoni2019monoloco}.

The challenge with deep heteroscedastic regression is that, while mean estimation is supervised, the covariance lacks direct supervision and needs to be inferred. This creates optimization challenges when the predicted covariance is incorrect. For instance, \citet{skafte2019reliable} highlights that an incorrectly predicted small variance effectively increases the learning rate, affecting optimization. Similarly, \citet{seitzer2022on} observes that poor convergence is often accompanied with a large predicted variance, which further affects convergence. 

Several recent methods aim to alter the negative log-likelihood to mitigate the impact of the predicted covariance in optimization. \citet{seitzer2022on} addresses this by proposing $\beta$-NLL, which scales the negative log-likelihood objective (Eq. \ref{eq:nll}) with the predicted variance for optimization: $\mathcal{L}_{\beta-\textrm{NLL}} = \lfloor \Var(\hat{Y} | X)^{\beta} \rfloor * \mathcal{L}_{\textrm{NLL}}$. This scaling aims to reduce the impact of the predicted variance in the training process. While simple and effective, $\beta$-NLL is not a result of a valid distribution, and the optimized values do not translate to the variance of a distribution.

The recent method of \citet{stirn2023faithful} proposes an alternative approach by scaling the gradients of the mean estimator with the predicted covariance. Effectively, the mean estimator is trained to minimize the mean squared error, and the covariance estimator is trained to minimize the negative log-likelihood. This involves conflicting assumptions; while the mean estimator assumes that the multivariate residual is uncorrelated, the covariance estimator is expected to recover correlations from this residual \cite{immer2023effective}.

Unlike previous works which regularize the (co-)variance, \citet{immer2023effective} uses the natural parameterization of the Gaussian: $n_1 = \frac{\mu}{\sigma^2}$ and $n_2 = \frac{-1}{2\sigma^2}$ for regression. Additionally, the method uses Bayesian techniques to regularize the network as well as obtain a posterior over the parameters. Similar to \citet{seitzer2022on}, the method assumes a diagonal covariance matrix. However, this assumption diminishes the main advantages of learning the covariance, such as correlation analysis, sampling from the target distribution, and updating our predictions conditioned on partial observations of the target.

In contrast to previous works which focus on regularization as a means to improve optimization, this paper focuses on improving the predicted covariance\textit{ within the negative log-likelihood formulation}. The drawback of $\Cov(\hat{Y} | X) = g_{\Theta}(\vx)$ being an arbitrary mapping from $\vx$ to a positive definite matrix is common to all the aforementioned approaches. This drawback is significant since, in the absence of supervision, $g_{\Theta}(\vx)$ can take on any value that minimizes the objective and does not necessarily represent the randomness of the prediction. Therefore, we propose a novel closed-form approximation for the predicted covariance and show that incorporating the gradient and curvature of the prediction better explains its randomness.

\section{Taylor Induced Covariance (TIC)}
Let us return to the prediction distribution $q(\hat{Y} | X \!=\! \vx)$ and ponder on a fundamental question: What is the randomness of a prediction $\hat{\vy}$ for a sample $\vx$? Intuitively, we quantify the covariance as a function of how quickly the predicted mean changes within a small radius of $\vx$. Larger derivatives imply a rapid change in $\hat{\vy}$, and as a result the model has a higher variance about its estimate. 

We therefore proceed by introducing a heuristic interpretation of the neighborhood, which allows us to take principled steps towards a closed-form approximation.

\subsection{$\epsilon\,$- Neighborhood}

For a continuously distributed random variable $X$, the probability of exactly observing $p(X \!=\! x)$ is zero. Instead, the standard approach (for example Sec. 2.4 in \citep{evans2004probability}) is to observe $X$ in the neighborhood of $x$: $X \in [x - \delta, x + \delta]$. The definition of this neighborhood is not rigid, allowing for a heuristic interpretation. For instance, we can represent this neighborhood stochastically: $X = x + \epsilon$. Here, $x$ is the observation, and $\epsilon$ is a random variable, which we set to be a zero-mean isotropic Gaussian distribution $p(\bm{\epsilon}) = \mathcal{N}(0, \sigma^2_{\epsilon}(\vx) \mI_m)$ for future analysis. 

The advantage of this heuristic is that it allows us to represent $\hat{y} = f_{\theta}(\vx + \epsilon)$ stochastically. While the variance of $\epsilon$
is unknown (we will later show that it is learnt), we assume heteroscedasticity, which allows us to represent neighborhoods of varying spatial extents for each $\vx$. We therefore model $\Cov( f_{\theta}(x + \epsilon))$, and continue by taking the second order Taylor polynomial of $f_{\theta}(\vx + \epsilon)$.

\subsection{Second Order Taylor Polynomial}

The second order Taylor polynomial introduces the notion of gradient and curvature in modeling the covariance, and quantifies the rate at which a function can change within a small neighborhood around $\vx$. We have
\begin{align}
    f_{\theta}(\vx + \epsilon) &= f_{\theta}(\vx) + \mJ(\vx) \epsilon^T + \frac{\vh}{2}\;, \nonumber \\ \text{where } \vh_i &= \epsilon \, \tH_i(\vx) \epsilon^T \,\,\,\forall i \in 1 \dots n   \;.
    \label{eq:variance_taylor}
\end{align}
Here, $\vx \in \mathbb{R}^m$ is the input, $f_{\theta}(\vx) \in \mathbb{R}^n$ represents the multivariate prediction, $\epsilon \in \mathbb{R}^m$ represents the neighborhood of $\vx$,  $\mJ(\vx) \in \mathbb{R}^{n \times m}$ corresponds to the Jacobian matrix, and $\tH(\vx) \in \mathbb{R}^{n \times m \times m}$ represents the Hessian tensor. We note that all the individual terms in Eq. \ref{eq:variance_taylor} are $n$-dimensional.

\subsection{Covariance Estimation}

\label{sec:cov}
The covariance of Eq. \ref{eq:variance_taylor}, $\Cov  f_{\theta}(\vx + \epsilon)$, \textit{with respect to the random variable $\epsilon$ is} 
\begin{equation}
\resizebox{0.9\columnwidth}{!}{$
\Cov (\mJ(\vx) \epsilon^T) + \Cov (\dfrac{\vh}{2})
+ 2 \,\bigg[ \Cov(\mJ(\vx) \epsilon^T, \dfrac{\vh}{2}) \bigg]\;.
$}
\label{eq:cov_defn}
\end{equation}
We obtain this since $f_{\theta}(\vx)$ is a constant with respect to $\epsilon$. Below, we evaluate the three terms individually.

\subsubsection{Estimating $\Cov(\mJ(\vx) \epsilon^T, \vh / 2)$}

We begin by noting that $\mJ(\vx) \epsilon^T$ and $\vh$ are $n$-dimensional vectors with elements $[\dots \mJ_{i}(\vx) \epsilon^T \dots]$ and $[\dots  \epsilon \, \tH_k(\vx) \epsilon^T \dots]$, respectively. The covariance between any two elements is given by
\begin{align}
& \Cov \big(\mJ_{i}(\vx) \epsilon^T, \epsilon \, \tH_k(\vx) \epsilon^T \big) \nonumber \\ 
&= \mathbb{E} \big(  \mJ_{i}(\vx) \epsilon^T \epsilon \, \tH_k(\vx) \epsilon^T \big) -  \mathbb{E} \big(  \mJ_{i}(\vx) \epsilon^T \big) \mathbb{E} \big( \epsilon \, \tH_k(\vx) \epsilon^T \big) \nonumber \\
&= 0 \;.
\label{eq:cov_taylor_zeromean}
\end{align}

\textbf{Odd and Even Functions.} We use the property of odd-even functions \citep{shynk2012probability}, which is based on symmetry and anti-symmetry of a function. Recall that an odd function is defined as $f(-t) = -f(t)$ and an even function as $f(-t) = f(t)$. Furthermore, the product of an odd and an even function is odd, and the product of two even functions is even. Finally, the integral of an odd function over its domain evaluates to zero.

We note that $\mJ_{i}(\vx) \epsilon^T = \sum_k \mJ_{i, k}(\vx) \epsilon^T_k$ is an odd function with respect to $\epsilon$. Furthermore, our design choice of $p(\bm{\epsilon}) = \mathcal{N}(0, \sigma^2_{\epsilon}(\vx) \mI_m)$ implies that $p(\bm{\epsilon})$ is an even function. The term $\mathbb{E} \big(  \mJ_{i}(\vx) \epsilon^T \big)$ can be written as $\int_{\epsilon}  \mJ_{i}(\vx) \epsilon^T p(\epsilon) \textrm{d} \epsilon $. This term represents the integral of a product of an odd and an even function, which evaluates to zero.

The quadratic term $\epsilon \, \tH_k(\vx) \epsilon^T$ can be written as $\sum_i\sum_j \tH^{(k)}_{i,j} \epsilon_i \epsilon_j$, which is an even function. Subsequently, $\mJ_{i}(\vx) \epsilon^T \epsilon \, \tH_k(\vx) \epsilon^T$  is a product of odd $ \mJ_{i}(\vx) \epsilon^T$ and even $\epsilon \, \tH_k(\vx) \epsilon^T$ terms. Finally, we can write $\mathbb{E} \big( \mJ_{i}(\vx) \epsilon^T \epsilon \, \tH_k(\vx) \epsilon^T \big)$ as $\int_{\epsilon}  \mJ_{i}(\vx) \epsilon^T   \epsilon \, \tH_k(\vx) \epsilon^Tp(\epsilon) \textrm{d} \epsilon $, which represents the integral of a product of odd, even, and even functions, which also evaluates to zero.

As a result, we get $\Cov \big( \mJ_{i}(\vx) \epsilon^T, \epsilon \, \tH_k(\vx) \epsilon^T \big)$ = 0 $\forall i, k$, implying that $\Cov(\mJ(\vx) \epsilon^T, \vh / 2) = 0$.

\subsubsection{Estimating $\Cov (\mJ(\vx) \epsilon^T)$ and $\Cov (\vh / 2)$}

Estimating $\Cov (\mJ(\vx) \epsilon^T)$ and $\Cov (\vh / 2)$ in Eq. \ref{eq:cov_defn} is easier since they follow a linear and quadratic form, respectively, with known solutions for isotropic Gaussian random variables (Eq. 375, 379 in~\citep{IMM2012-03274}). Specifically, we have
\begin{align}
&\Cov(\mJ(\vx) \,\, \epsilon^T) = k_1(x) \mJ(\vx) \mJ(\vx)^T \nonumber \\
&\Cov(\vh / 2)_{i, j} = k_2(x) \,\,\texttt{Trace}\,\,(\tH_{i, :, :}(\vx) \,\, \tH_{j, :, :}(\vx))\;.
\label{eq:lin_quad_form}
\end{align}
Since we do not know the variance of the $\epsilon$ and its transformation for each $\vx$, we define them through positive quantities $k_1(\vx)$ and $k_2(\vx)$, which are optimized by the covariance estimator $g_{\Theta}(\vx)$. We also note that both $\Cov(\mJ(\vx) \, \epsilon^T)$ and $\Cov(\vh / 2)$ have dimensions $n \times n$. 

Finally, we obtain the solution for Eq. \ref{eq:cov_defn} by substituting Eq.~\ref{eq:lin_quad_form} and Eq.~\ref{eq:cov_taylor_zeromean} into it, which yields
\begin{align}
\Cov  f_{\theta}(\vx + \epsilon) &= k_1(\vx) \mJ(\vx) \mJ(\vx)^T + \mathcal{H} \nonumber \\
\text{where } \mathcal{H}_{i, j} &= k_2(\vx) \,\texttt{Trace}\,(\tH_{i, :, :}(\vx) \, \tH_{j, :, :}(\vx))\;.
\label{eq:prefinal}
\end{align}

\subsection{Formulation}

We defined the covariance through $\epsilon$, the neighborhood random variable which allows us to capture the gradient and curvature of $f_{\theta}(x + \epsilon)$. However, the target $y$ could have stochasticity that does not depend upon the neighborhood. We take as an example the function $y = c + \mathcal{N}(0, \Sigma(x))$. Here, the stochasticity of $y$ is independent of the neighborhood $\epsilon$. To address scenarios such as these, we introduce a new random variable $\varepsilon \sim \mathcal{N}(0, \Sigma(x))$, which is heteroscedastic and is independent of $f_{\theta}(x + \epsilon)$. Indeed, $\varepsilon$ does not depend upon the gradient or curvature of $f_{\theta}$. Subsequently, we can write $\hat{y} = f_{\theta}(x + \epsilon) + \varepsilon$. Since $\varepsilon$ is independent of $f_{\theta}(x + \epsilon)$, we can write the covariance as the sum of $\textrm{Cov} f_{\theta}(x + \epsilon)$ and $\Sigma(x)$. This is possible because the sum of two independent Gaussians results in a Gaussian with the means and covariances summed. We therefore model the covariance through the gradient and curvature as well as account for the inherent stochasticity of the samples.

\begin{algorithm}
\small
\DontPrintSemicolon
\KwInput{$\vx$: Input sample}
\KwInput{$f_{\theta}$: Mean estimator}
\KwOutput{$\Cov(\hat{Y} | X)$: Covariance prediction}
\vspace{0.2cm}

\tcp{Parallelized using vmap}

\vspace{0.1cm}
$\mJ(\vx) = \texttt{get\_jacobian\_wrt\_x}(f_{\theta}(x))$

$\tH(\vx) = \texttt{get\_hessian\_wrt\_x}(f_{\theta}(x))$

$k_1(\vx), k_2(\vx), k_3(\vx) = g_{\Theta}(\vx)$

\vspace{0.2cm}

\tcp{Jacobian term}
\tcp{J.shape = (out\_dims $\times$ in\_dims)}
jacobian = $k_1(\vx) \mJ(\vx) \mJ(\vx)^T$

\vspace{0.2cm}

\tcp{Hessian term}
\tcp{H.shape = (out\_dims $\times$ in\_dims $\times$ in\_dims)}

$\mathcal{H}_{i, j} = k_2(\vx) \,\texttt{Trace}\,(\tH_{i, :, :}(\vx) \, \tH_{j, :, :}(\vx))$

hessian = $\mathcal{H}$

\vspace{0.2cm}

\tcp{Independent term}

independent = $k_3(\vx)$

\vspace{0.2cm}

\tcp{Taylor Induced Covariance}
TIC = jacobian + hessian + independent

\vspace{0.2cm}
\tcp{Train using negative loglikelihood}
\Return TIC
\caption{\textbf{\textit{Taylor Induced Covariance}}}
\label{algorithm:tic}
\end{algorithm}

We learn the covariance of $\varepsilon$ through $k_3(\vx) \in \mathbb{R}^{n \times n}$, a learnable positive definite matrix which is optimized via the covariance estimator $g_{\Theta}(x)$. The final expression for TIC is
\begin{align}
    \Cov(\hat{Y} | X \!=\!x) &\approx k_1(\vx) \mJ(\vx) \mJ(\vx)^T + \mathcal{H} + k_3(\vx)\;.
    \label{eq:final}
\end{align}
The covariance estimator $g_{\Theta}(x)$ predicts $k_1(\vx), k_2(\vx)$ and $k_3(\vx)$, where $k_1(\vx), k_2(\vx)$ are positive scalars. We enforce $k_3(x)$ to be positive definite by predicting an unconstrained matrix and multiplying it with its transpose, similarly to previous work. The covariance estimator is jointly optimized with the mean and is trained to minimize the negative log-likelihood by substituting Eq. \ref{eq:final} into Eq. \ref{eq:nll}.

\subsection{Discussion}

At first, the use of the Hessian in TIC resembles its use in optimization \citep{gilmer2022a, kingma2014adam}. The Cramer-Rao bound \citep{ly2017tutorial} links the variance of a parametric estimator with its inverse Fisher information. However, the Fisher information computes the Hessian with respect to the \textit{parameters}, measuring its sensitivity over all samples. By contrast, the Hessian in our formulation is computed with respect to the input, allowing us to model heteroscedasticity.

We incorporate TIC within the negative log-likelihood formulation and do not employ covariance specific regularization. Similar to previous works \citep{kendall2017uncertainties, skafte2019reliable, seitzer2022on, stirn2023faithful, immer2023effective}, our method is an approximation without theoretical guarantees. This approximation results from a heuristic interpretation of the neigborhood, as well as the use of the second order taylor polynomial.  However, our experimental evaluations show that TIC provides accurate covariance estimates and works well in practice.

\textbf{Limitations. } The computational complexity in TIC arises from computing the Hessian. While determining this for a general network architecture is non-trivial, computing the Hessian for a function that maps $x \in \mathbb{R}^m$ to $y \in \mathbb{R}^n$ has a complexity of $O(nm^3)$ \citep{yao2020pyhessian}, which is large. There are multiple possible ways to mitigate this in practice. The simplest approach would be to use parallelization, which we provide in our code. The second approach would be to use a smaller, proxy model in place of a large model (which could be retained for mean estimation). This smaller model could be trained through a student-teacher setup using techniques from knowledge distillation \citep{gou2021knowledge}. The reduced parameter count would decrease the computational requirements of the Hessian. An interesting direction for future research would be to find useful approximations of the Hessian with respect to the input, similarly to research in optimization which approximates the Hessian with respect to the parameters.

\section{Task Agnostic Correlations (TAC)}

\begin{figure}
\centering
    \includegraphics[trim=0.0cm 0.0cm 1.0cm 2.0cm, width=0.8\columnwidth]{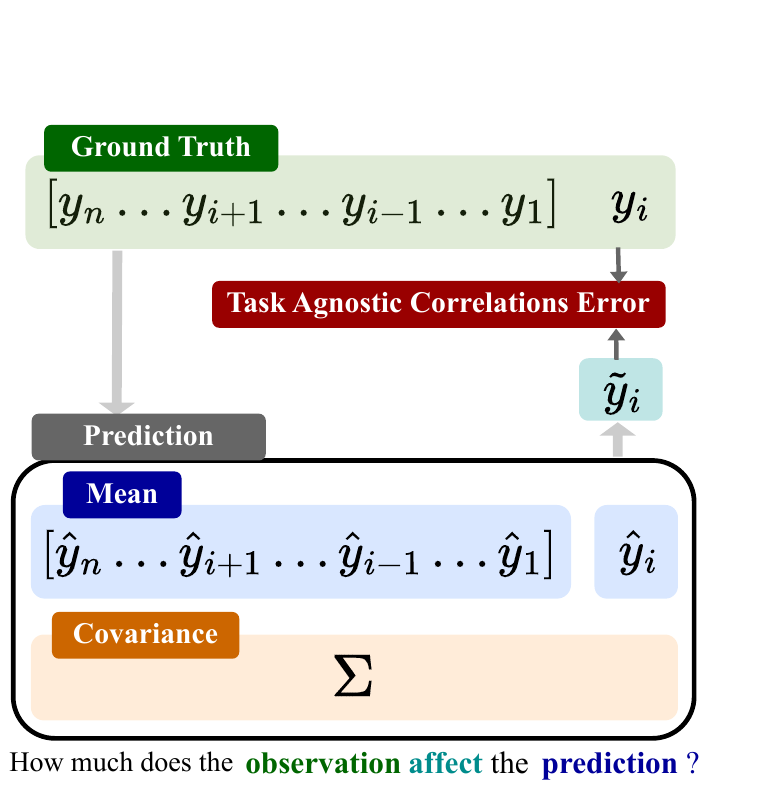}
    \caption{\textit{Task Agnostic Correlations (TAC).} We propose the TAC metric for covariance evaluation. Given the ground truth $\vy$, predicted mean $\hat{\vy}$ and covariance $\mSigma$, TAC quantifies the improvement in the predicted mean given partial observations of the ground truth. TAC uses conditioning of the normal distribution 
    to directly assess the covariance.}
    \label{fig:tac}
\end{figure}

\begin{figure*}
    \centering
    \begin{subfigure}{\textwidth}
        \includegraphics[width=\textwidth]{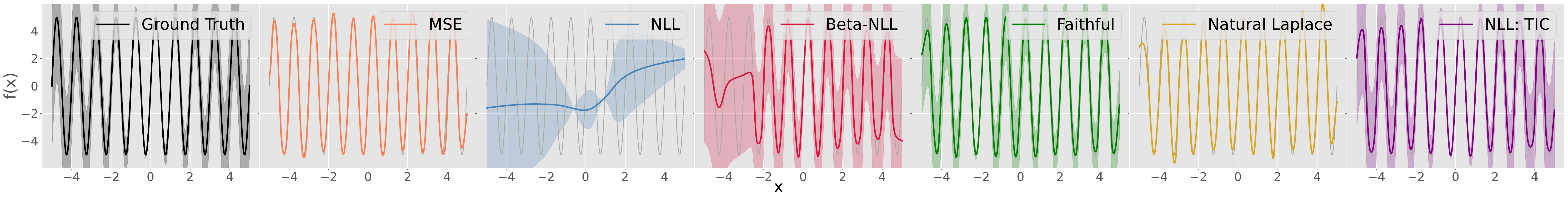}
        \vspace*{-7mm}
        \caption{Sinusoidal with  $y = 5 \textrm{ sin } (2 \pi x)$ and $\sigma(x) =  |x|$}
    \end{subfigure}
    \centering
    \begin{subfigure}{\textwidth}
        \includegraphics[width=\textwidth]{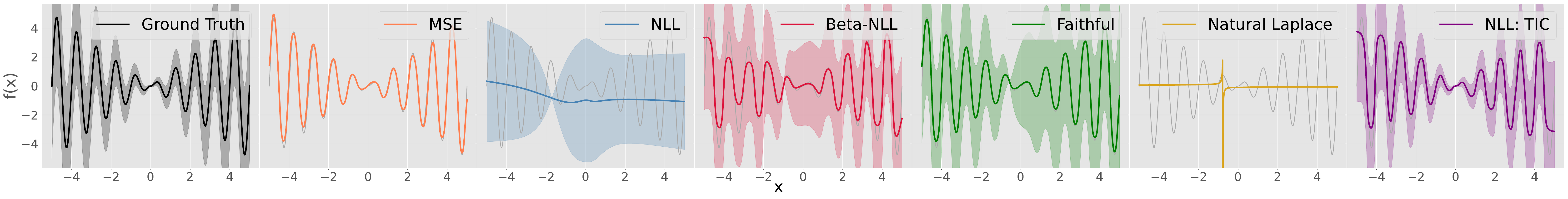}
        \vspace*{-7mm}
        \caption{Sinusoidal with  $y = |x| \textrm{ sin } (2 \pi x)$ and $\sigma(x) =  |x|$}
    \end{subfigure}
    \begin{subfigure}{\textwidth}
        \includegraphics[width=\textwidth]{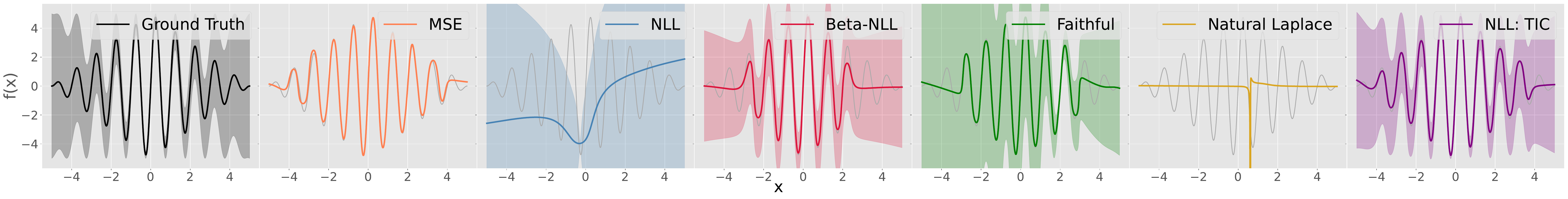}
        \vspace*{-7mm}
        \caption{Sinusoidal with  $y = (5 - |x|) \textrm{ sin } (2 \pi x)$ and $\sigma(x) =  |x|$}
    \end{subfigure}
    \caption{\textit{Univariate. } We perform experiments on three different sinusoidals, showing that incorporating the gradient and curvature of the predicted mean results in accurate variance estimation. The TIC parameterization also results in an improved convergence of the negative log-likelihood.}
    \label{fig:univariate}
\end{figure*}

\begin{algorithm}
\small
\DontPrintSemicolon
\KwInput{\textcolor{OliveGreen}{$\vy$}: Ground truth, \hspace{0.25cm} \textcolor{BlueViolet}{$\hat{\vy}$}: Target prediction}
\KwInput{\textcolor{Bittersweet}{$\Cov(\hat{Y} | X)$}: Covariance prediction}
\KwOutput{\textcolor{Mahogany}{TAC} error}
\vspace{0.2cm}

dimensions = \texttt{get\_dimensions}($\hat{\vy}$) 

\textcolor{Mahogany}{TAC} = \texttt{zeros}(shape=dimensions)

\vspace{0.2cm}

\For{i in dimensions}    
{%
    \vspace{0.1cm}
    \tcp{Observe all but one dimension}
    obs\_dim = \texttt{set}(dimensions) - \texttt{set}(i)
                    
    hidden\_dim = i

    \vspace{0.1cm}
    
    \tcp{Conditioning the normal distribution}
    \textcolor{Bittersweet}{$\mSigma_{22}$} = $\Cov(\hat{Y} | X)$[obs\_dim, obs\_dim]
                    
    \textcolor{Bittersweet}{$\mSigma_{12}$} = $\Cov(\hat{Y} | X)$[hidden\_dim, obs\_dim]
                    
    \textcolor{Cerulean}{$\tilde{\vy}$} = \textcolor{BlueViolet}{$\hat{\vy}$[hidden\_dim]} + (\textcolor{Bittersweet}{$\mSigma_{12} \mSigma_{22}^{-1}$} ($\textcolor{OliveGreen}{\vy[\textrm{obs\_dim}]} - \textcolor{BlueViolet}{\hat{\vy}[\textrm{obs\_dim}]}$))
    
    \vspace{0.2cm}

    \tcp{Error between updated and true value}
    \textcolor{Mahogany}{TAC [i]} = $|\,\textcolor{Cerulean}{\tilde{\vy}} - \textcolor{OliveGreen}{\vy[\textrm{hidden\_dim}]\,|}$
}%
\vspace{0.2cm}
\Return TAC.\texttt{mean()}
\caption{\textbf{\textit{Task Agnostic Correlations}}}
\label{algorithm:tac}
\end{algorithm}

How can we evaluate covariance estimation in the absence of ground-truth annotation? Existing methods \citep{kendall2017uncertainties, seitzer2022on, stirn2023faithful} use metrics such as likelihood scores and the mean squared error for evaluation. However, these methods are skewed towards learning the mean; a perfect mean estimator $f_{\theta}(x)$ would result in zero mean squared error, while log-likelihood scores put greater emphasis on the determinant of the covariance and do not directly assess correlations. Other metrics such as the Conditional Marginal Likelihood (CML) \citep{lotfi2022bayesian} are a measure of generalization. Therefore, we argue for the use of a much more direct method to assess the covariance. Specifically, we reason that the goal of estimating the covariance is to encode the relation between the target variables. \textit{Therefore, partially observing a set of correlated targets should improve the prediction of the hidden targets since by definition the covariance encodes this correlation.} As an example, if $P$ and $Q$ are correlated, then observing $P$ should improve our estimate of $Q$. Hence, we propose a new metric that evaluates the accuracy of correlations, which we call the \textit{Task Agnostic Correlations},  (Figure \ref{fig:tac}).

\subsection{Algorithm}

Formally, given  an $n$-dimensional target prediction $\hat{\vy}$, the ground truth $\vy$, and the predicted covariance $\Cov(\hat{Y} | X \!\! = \!\! \vx)$, we define the TAC error as $\sum_i |y_i - \Tilde{y}_i| / n$, where $\Tilde{y}_i$ is the updated mean obtained after conditioning $\mathcal{N}(\tilde{y}_{i}, \Cov(\hat{Y} | X) \,\, | \,\, \vy_{j \neq i}, \vx)$. For each prediction $\hat{y}_i$, we obtain a revised estimate $\tilde{y}_i$ by conditioning it over the ground truth of the remaining variables $\vy_{i \neq j}$. We measure the absolute error of this revised estimate against the ground truth of the unobserved variable and repeat for all $i$. An accurate estimate of $\Cov(\hat{Y} | X \!\! = \!\! \vx)$ will decrease the error whereas an incorrect estimate will cause an increase. We describe this in Algorithm \ref{algorithm:tac}.

\subsection{Discussion}

This evaluation bears resemblance with \textit{leave-one-out}, where we observe one $\tilde{y}_i$ given other observations $\vy_{j \neq i}$. While \textit{leave-one-out} can be generalized to \textit{leave-k-out}, we do not observe any change in the evaluation trend. A method having lower \textit{leave-one-out} also has a lower \textit{leave-k-out} error. Moreover, \textit{leave-k-out} requires taking $\binom{n}{k}$ combinations, which is significantly higher than taking $n$ combinations in \textit{leave-one-out}. This motivates the use of the \textit{leave-one-out}.

We highlight that this metric is agnostic of downstream tasks involving covariance estimation. \textit{We also note that TAC and the log-likelihood are complementary: while log-likelihood is a measure of optimization, TAC is a measure of accuracy of the learnt correlations.} Hence, we use TAC as an additional metric for all multivariate experiments.

\section{Experiments}

The goal of this paper is to improve covariance estimation in deep heteroscedastic regression. Therefore, we specifically focus on multivariate \textit{outputs}, and readdress several existing experimental designs. Our synthetic experiments consist of learning a univariate sinusoidal and a multivariate distribution. We conduct our real-world experiments on the UCI regression repository and the MPII, LSP 2D human pose estimation datasets. 

Our baselines consist of different (co-)variance models in deep heteroscedastic regression. These include the negative log-likelihood \cite{dorta2018structured}, and its variants: $\beta$-NLL \citep{seitzer2022on}, Faithful Heteroscedastic Regression \citep{stirn2023faithful}, and Natural Laplace \cite{immer2023effective} (univariate). We refer to the diagonal covariance \citep{kendall2017uncertainties} and the TIC formulation as \textit{NLL-Diagonal} and \textit{NLL-TIC} since they are optimized using the negative log-likelihood. We take care to provide a fair comparison; all methods are randomly initialized with the same mean and covariance estimators, with each method having its own learning rate scheduler. Furthermore, the batching and ordering of samples is the same for all methods. We train our method and all baselines for 100 epochs using a learning rate scheduler which reduces the learning rate on plateau. Unless specified, we use simple fully connected layers with batch normalization as our network architecture.

Since covariance estimation lacks direct supervision, we do not make training and evaluation splits of the dataset to increase the number of samples. While this may seem questionable, we reason that the covariance is a measure of correlation as well as variance. If too few samples are provided for training then the resulting covariance is nearly singular. Moreover, existing work \cite{skafte2019reliable, seitzer2022on, stirn2023faithful} show that the negative log-likelihood is prone to sub-optimal convergence and does not overfit the training samples. Finally, our evaluation remains fair since our experimental methodology is the same for all.

\subsection{Synthetic Data}

\textit{\textbf{Univariate}.} We repeat the experiments of \citet{seitzer2022on} with a major revision. First, we introduce heteroscedasticity and substantially increase the variance of the samples. Second, we simulate different sinusoidals having constant and varying amplitudes. We draw 50,000 samples and train a fully-connected network with batch normalization \cite{ioffe} for 100 epochs. Our results are shown in Figure \ref{fig:univariate}. We observe that in the absence of direct supervision, the negative log-likelihood incorrectly overestimates the variance since it does not represent the randomness of the predicted mean. Furthermore, both $\beta$-NLL and Faithful are susceptible to incorrect variance predictions because the methods regularize the variance, which compromises on variance fits. While Natural Laplace fits the constant amplitude sinusoidal, the method results in unstable optimization for sinusoidals of varying amplitude.

\begin{figure}
    \centering \includegraphics[width=0.8\columnwidth]{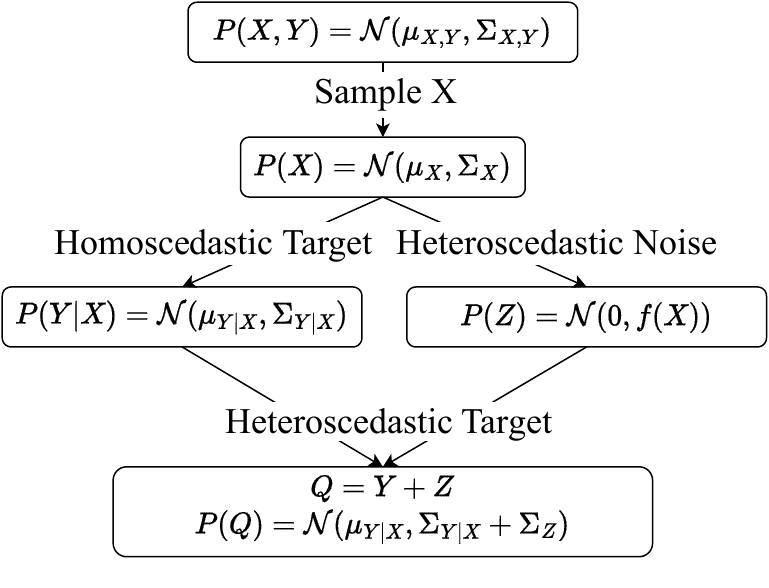}
    \caption{\textit{Multivariate Schematic. }We present a simple method to simulate heteroscedastic data. We first randomly sample the input $\vx$, which in turn is used to sample an initial target $\vy$. We then add sample-dependent noise $\vz$, giving us the target $\vq$, which the network is required to learn.}
    \label{fig:multivariate}
\end{figure}

\textit{\textbf{Multivariate}}. We propose a new experiment for multivariate analysis to study heteroscedastic covariance. We let $X, Y$ be jointly distributed and sample $\vx$ from this distribution. Subsequently, we sample $\vy$ conditioned on $\vx$. To simulate heteroscedasticity, we draw samples from $Z$, a zero mean random variable whose covariance $\Sigma_Z = \text{diag}(\sqrt{|\vx|})$ depends on $\vx$. Since $Y$ and $Z$ are independent given $X$, their sum also satisfies the normal distribution $Q | X \sim \mathcal{N}(\mu_{Y | X}, \Sigma_{Y | X} + \Sigma_{Z | X})$. Therefore, the goal of this experiment is to model the mean and the heteroscedastic covariance of $Q|X$ given samples $(\vx, \vq)$. The schematic for our experimental design is shown in Fig. \ref{fig:multivariate}. 

\begin{table}
\caption{\textit{Multivariate Results. }We compare the log-likelihood value for all methods. We skip NLL-Diagonal and $\beta$-NLL, both of which have very low likelihoods since they assume a diagonal covariance.}
\label{tab:ll_mv}
\renewcommand{\arraystretch}{1.0}
\centering
\resizebox{\linewidth}{!}{%
\begin{tabular}{lccccccccc}
    \toprule
    Method & Dim: 4 & 6 & 8 & 10 & 12  & 14 & 16 & 18 & 20\\
    \midrule
    MSE &  -10.1 & -16.4 &  -23.1 &  -30.9 & -36.1  & -41.6 & -49.2 &  -53.2 &  -66.6\\
    NLL &  -8.2 &  -14.9 &  -19.9 & -26.6 & -34.2 & -42.7 & -46.6 & -60.9 & -67.2\\
    Faithful & -8.7 & -14.7 & -20.3 & -27.3 & -32.4 & -40.2 & -48.6 & -55.4 & -69.0\\
    \midrule
    \textbf{NLL-TIC} &  \textbf{-7.6} & \textbf{-11.7} & \textbf{-15.8} & \textbf{-19.9} & \textbf{-23.3} & \textbf{-26.8} & \textbf{-30.3} & \textbf{-34.2} &  \textbf{-39.7}\\
    \bottomrule
\end{tabular}
}
\end{table}

\begin{figure}
    \centering \includegraphics[width=\columnwidth]{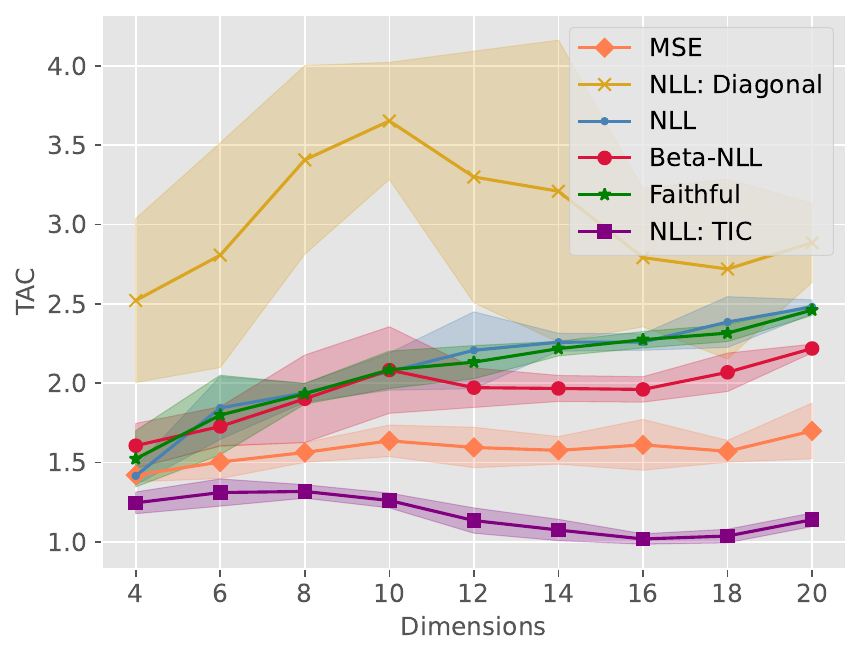}
    \caption{\textit{Multivariate Results.} We plot the Task Agnostic Correlations (TAC) metric mean and standard deviation for all methods from dimensions 4 to 20. The gap between TIC and the baselines widens as the dimensionality increases.}
    \label{fig:tac_mv}
\end{figure}
We vary the dimensionality of $\vx$ and $\vq$ from 4 to 20 in steps of 2, and report the mean and standard deviation over ten trials for each dimension. Depending on the dimensionality, we draw from 4000 up to 20000 samples and report our results using TAC (Fig. \ref{fig:tac_mv}) and the log-likelihood (Table \ref{tab:ll_mv}). We observe two trends in Fig. \ref{fig:tac_mv}: First, as the dimensionality of the samples increases, the gap between TIC and the other methods widens. This is because, with increasing dimensionality, the number of free parameters to estimate in the covariance matrix grows quadratically. An increase in parameters typically requires a non-linear growth in the number of samples for robust fitting. As a result, the difficulty of  mapping the input to a positive definite matrix increases with dimensionality. Second, we note that TIC allows for better convergence in comparison to the naive parameterization of the covariance in NLL.

\subsection{UCI Regression}

\begin{table*}
\caption{\textit{UCI Regression}. We perform ten trials over all the datasets and report the TAC error and the log-likelihood. TIC outperforms all the baselines on ten out of twelve datasets in terms of TAC error, and outperforming on all but one dataset in terms of the likelihood error. Additionally, the TIC parameterization (NLL-TIC) results in improved convergence of the negative log-likelihood.}
\centering
\renewcommand{\arraystretch}{1.0}
\begin{subtable}[t]{\textwidth}
    \centering
    \caption{Task Agnostic Correlations (TAC) Metric}
    \resizebox{\linewidth}{!}{%
    \begin{tabular}{lcccccccccccc}
        \toprule
        Method & Abalone & Air & Appliances & Concrete & Electrical & Energy & Turbine & Naval & Parkinson & Power & Red Wine & White Wine \\
        \midrule
        MSE & 2.54 & 4.31 & 1.79 & 6.15 & 7.91 & 4.40 & 4.74 & 0.56 & 2.32 & 6.01 & 5.97 & 6.32 \\
        NLL-Diagonal & 5.49 & 8.03 & 11.71 & 7.86 & 10.06 & 7.12 & 7.07 & 5.01 & 8.56 & 8.16 & 7.96 & 8.44 \\
        NLL & 3.28 & 3.42 & 2.41 & 4.16 & 7.14 & 5.10 & 3.40 & 0.25 & 1.86 & 6.22 & 5.81 & 7.26 \\
        $\beta$-NLL  & 2.85 & 5.67 & 4.89 & 7.21 & 8.41 & 6.17 & 5.03 & 1.06 & 5.48 & 6.73 & 6.96 & 7.08 \\
        Faithful & 2.96 & 3.27 & 1.79 & 3.93 & 7.36 & 2.90 & 3.29 & \textbf{0.20} & \textbf{1.68} & 5.81 & 5.74 & 6.89 \\
        \midrule
        \textbf{NLL-TIC} & \textbf{1.83} & \textbf{2.27} & \textbf{1.39} & \textbf{2.82} & \textbf{4.89} & \textbf{2.34} & \textbf{2.40} & 0.28 & 2.54 & \textbf{3.87} & \textbf{4.05} & \textbf{4.60} \\
        \bottomrule
        \vspace{1mm}
    \end{tabular}
    }
    \label{subtab:first}
\end{subtable}
\begin{subtable}[t]{\textwidth}
    \centering
    \caption{Log Likelihood Metric. We skip NLL-Diagonal and $\beta$-NLL which have very low likelihoods since the methods assume diagonal covariance. We remind the reader that the datasets are adapted for covariance estimation}
    \resizebox{\linewidth}{!}{%
    \begin{tabular}{lcccccccccccc}
        \toprule
        Method & Abalone & Air & Appliances & Concrete & Electrical & Energy & Turbine & Naval & Parkinson & Power & Red Wine & White Wine \\
        \midrule
        MSE & -60.7 & -231.5 & -99.6 & -238.3 & -494.6 & -169.6 & -230.8 & -20.9 & -154.0 & -295.6 & -305.8 & -338.15 \\
        NLL & $-8.5 \times 10^3$ & -53.32 & -84.5 & -83.6 & -57.9 & -55.8 & -27.1 & 4.1 & $-1.5 \times 10^3$ & -34.2 & -236.0 & -206.0 \\
        Faithful & $-9.4 \times 10^3$ & -52.1 & -55.4 & -80.6 & -57.3 & -30.8 & -26.1 & \textbf{7.5} & $-1.2 \times 10^3$ & -33.9 & -434.4 & -250.9 \\
        \midrule
        \textbf{NLL-TIC} & \textbf{-13.4} & \textbf{-29.3} & \textbf{-42.45} & \textbf{-22.2} & \textbf{-35.8} & \textbf{-19.1} & \textbf{-22.9} & -10.3 & \textbf{-63.0} & \textbf{-27.1} & \textbf{-30.63} & \textbf{-30.1} \\
        \bottomrule
    \end{tabular}
    }
    \label{subtab:second}
\end{subtable}
\label{tab:overall}
\end{table*}

We perform our analysis on twelve multivariate UCI regression \cite{Dua:2019} datasets, which have been used in previous work on negative log-likelihood \cite{stirn2023faithful, seitzer2022on}. Nevertheless, our goal of studying covariance estimation in deep heteroscedastic regression requires us to use a different pre-processing, as many of the datasets have univariate or low-dimensional targets.

Specifically, for each dataset we randomly allocate 25\% of the features as input and the remaining 75\%  features as multivariate targets at run-time. Some combinations of input variables may fare poorly at predicting the target variables. However, this is an interesting challenge for the covariance estimator, which needs to learn the underlying correlations even in unfavorable circumstances. Moreover, random splitting also allows our experiments to remain unbiased as we do not control the split of variables at any instant. For all datasets, we standardize our variables with zero mean and a variance of ten (which yields better convergence for all methods). We perform 10 trials for each dataset and report the TAC error and likelihood in  Table \ref{tab:overall}.

While TIC outperforms the baselines on a majority of the datasets, we particularly focus on the Naval dataset which highlights a limitation of the TIC parameterization. We observe that TIC may not be suitable if all samples have a low degree of variance. A low degree of variance (as indicated by the likelihood) results in accurate mean fits, which implies that small gradients are being backpropagated, and in turn affecting the TIC parameterization. However, we argue that datasets with a small degree of variance may not benefit from heteroscedastic modelling.

\subsection{2D Human Pose Estimation}

\begin{figure*}[h]
    \centering
    \includegraphics[width=\linewidth]{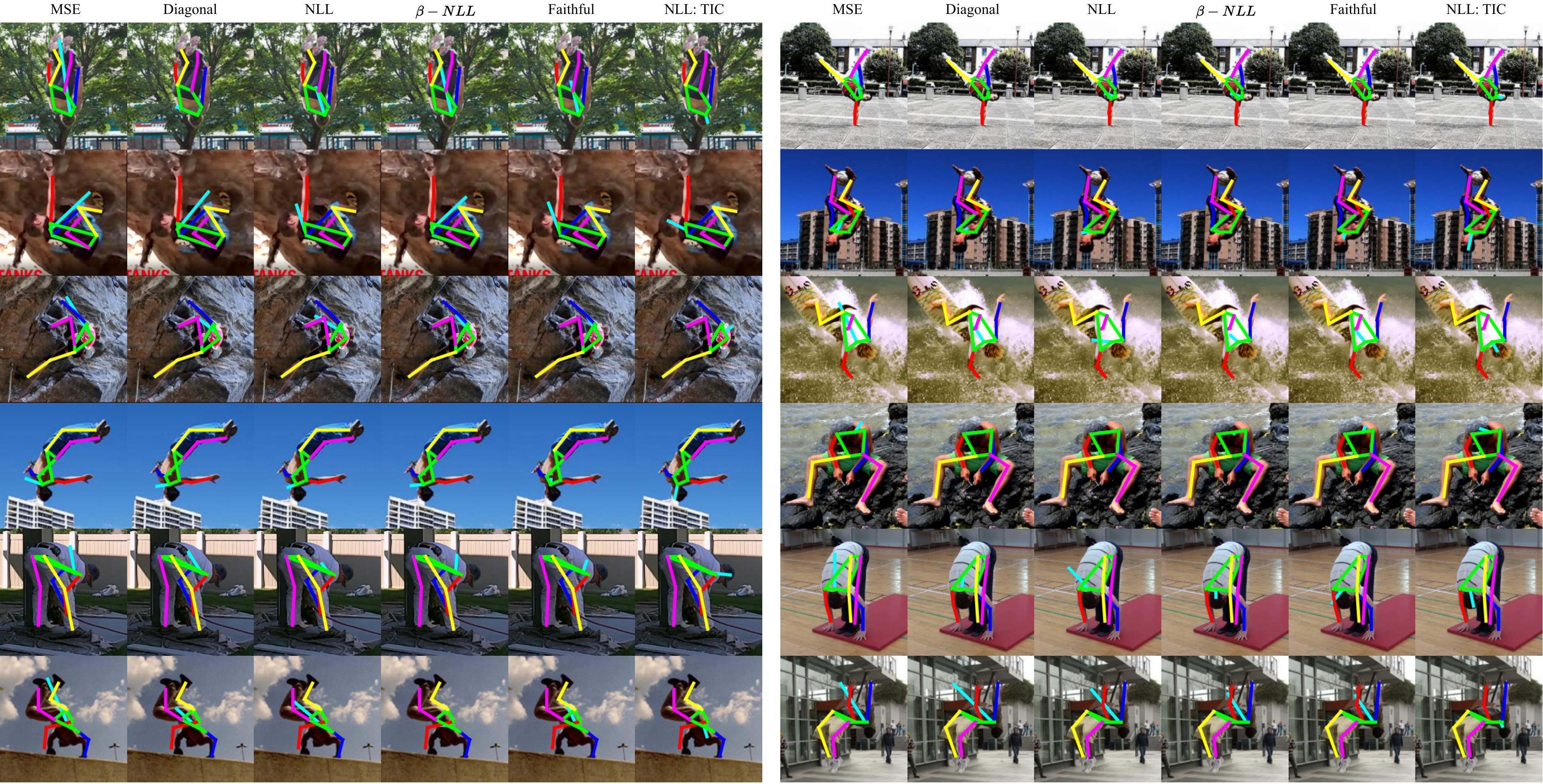}
    \caption{\textit{Human Pose Visualization}. We show that the Taylor Induced Covariance (TIC) parameterization results in a more accurate pose estimation for complex poses. As an example, we visualize the updated prediction for the head conditioned on observing the ground truth for the remaining joints. We show that TIC accurately predicts the location for the head for complex poses in comparison to all other methods.}
    \label{fig:hp_viz}
\end{figure*}
\begin{table*}
  \caption{\textit{Human Pose Results - ViTPose architecture}. We report the TAC error for each joint along with the average across all joints. Additionally, we report the likelihood score for all methods. We show that NLL-TIC outperforms baselines across all joints and successfully scales to convolutional and transformer based architectures.}
  \label{tab:humanpose}
  \renewcommand{\arraystretch}{1.0}
  \centering
  \resizebox{\linewidth}{!}{%
  \begin{tabular}{lcccccccccccccccccr}
    \toprule
    Method & head & neck & lsho & lelb & lwri  & rsho & relb & rwri & lhip & lknee &  lankl & rhip & rknee & rankl & \textbf{Avg: TAC} & \textbf{Avg: LL}\\
    \midrule
    MSE & 6.14 & 7.12 & 7.05 & 8.60 & 10.56 & 6.78 & 8.33 & 10.35 & 7.67 & 7.90 & 9.69 & 7.40 & 7.82 & 9.72 & 8.22 $\pm$ 0.05 & -973.7 $\pm$ 8.6\\
    
    NLL-Diagonal & 14.88 & 12.33 & 12.38 & 12.25 & 13.87 & 11.36 & 11.39 & 13.54 & 10.42 & 11.49 & 17.84 & 9.84 & 11.46 & 18.28 & 12.95 $\pm$ 1.36 & -204.5 $\pm$ 177.0\\
    
    NLL & 4.97 & 5.76 & 4.86 & 4.58 & 6.62 & 4.48 & 4.36 & 6.59 & 5.97 & 5.80 & 7.88 & 5.78 & 5.68 & 7.81 & 5.80 $\pm$ 0.07 & -91.61 $\pm$ 1.26\\
    
    $\beta$-NLL & 12.63 & 11.22 & 11.63 & 12.06 & 13.95 & 10.45 & 11.21 & 13.63 & 10.84 & 11.45 & 16.23 & 10.02 & 11.09 & 15.97 & 12.31 $\pm$ 0.31 & -4.2e3 $\pm$ 1.6e3\\

    Faithful & 5.25 & 5.86 & 4.97 & 4.68 & 6.77 & 4.60 & 4.45 & 6.75 & 6.10 & 5.98 & 7.90 & 5.94 & 5.82 & 7.89 & 5.93 $\pm$ 0.03 & -91.77 $\pm$ 0.11\\
    \midrule
    \textbf{NLL-TIC} & \textbf{3.97} & \textbf{5.38} & \textbf{4.47} & \textbf{4.29} & \textbf{6.06} & \textbf{4.12} & \textbf{4.08} & \textbf{5.89} & \textbf{5.45} & \textbf{5.24} & \textbf{7.03} & \textbf{5.25} & \textbf{5.09} & \textbf{6.97} & \textbf{5.23} $\pm$ \textbf{0.03} & \textbf{-80.31} $\pm$ \textbf{0.39}\\
    \bottomrule
  \end{tabular}
  }
\end{table*}

We introduce experiments on human pose estimation since the human pose is an organised collection of points and is naturally suited for correlation analysis \cite{shukla2022vl4pose}. Moreover, popular human pose architectures \citep{hg, hrnet, kreiss2019pifpaf, kreiss2021openpifpaf, xu2022vitpose} are either convolutional or transformer based, presenting a viable challenge to modelling the Taylor Induced Covariance. This is because TIC assumes vector inputs $\vx \in \mathbb{R}^m$ and predictions $\hat{\vy} \in \mathbb{R}^n$, whereas popular architectures rely on input images $\tX \in \mathbb{R}^{C \times H \times W}$ and output heatmaps $\hat{\tY} \in \mathbb{R}^{\# \text{joints} \times 64 \times 64}$. 

We therefore perform experiments on two architectures: the Stacked Hourglass \citep{hg} and ViTPose \citep{xu2022vitpose}. The Stacked Hourglass is a popular method which extends the convolutional U-Net \cite{unet} architecture to predict heatmaps for human pose estimation. ViTPose is a recent state-of-the-art architecture which extends vision transformers \cite{dosovitskiy2021an} to the task of human pose estimation. 

For both architectures, we use soft-argmax \citep{li2021hybrik, li2021localization} to reduce the heatmap of shape $\hat{\tY} \in \mathbb{R}^{\# \text{joints} \times 64 \times 64}$ to a vector of shape $\mathbb{R}^{\#\text{joints} * 2}$. Next, we recursively call the hourglass module until we obtain a one-dimensional vector encoding \cite{shukla2022bayesian} for the image, which serves as the input to the covariance estimator. For ViTPose, we obtain vector embeddings from a simple residual connection involving a one-dimensional downscaling and upscaling of the features predicted by the backbone network. 

We use popular single person datasets: MPII \citep{mpii} and LSP/LSPET \cite{lsp, lspet}, with the latter emphasizing on poses involving sports. We perform our analysis by merging the MPII and LSP-LSPET datasets to increase the number of samples. We train the pose estimator using the Adam optimizer with a 'ReduceLROnPlateau' learning rate scheduler for 100 epochs with the learning rate set to 1e-3. We use two augmentations: Shift+Scale+Rotate and horizontal flip. We refer the reader to the code for implementation details.

In addition to the likelihood, we continue to use TAC as our metric since for single person estimation, the scale of the person is fixed. Hence, TAC is highly correlated with PCKh/PCK, the preferred metric for multi-person multi-scale pose estimation. We perform five trials and report our results for the \textit{ViTPose} backbone in Table \ref{tab:humanpose}. We report results on the Stacked Hourglass backbone in the appendix. Our experiments show that TIC outperforms all baselines, especially on challenging joints.

\section{Conclusion}
We improved covariance estimation in deep heteroscedastic regression through two contributions. With the Taylor Induced Covariance (TIC), we parameterize the predicted covariance to capture the randomness of the predicted mean through its gradient and curvature. With the Task Agnostic Correlations (TAC) metric, we have proposed a novel metric for covariance evaluation by leveraging conditioning of the normal distribution to quantify the accuracy of learnt correlations. Our extensive experiments across multiple tasks have shown that, not only does TIC outperform the state of the art in learning the covariance, it also facilitates an improved convergence of the negative log-likelihood.


\section*{Acknowledgements}

We thank the reviewers, for their valuable comments, insights as well as participation in the rebuttal period. We also thank \href{https://people.epfl.ch/reyhaneh.hosseininejad}{Reyhaneh Hosseininejad} for her help in preparing the paper and for her insights.

This research is funded by the Swiss National Science Foundation (SNSF) through the project \textit{Narratives from the Long Tail: Transforming Access to Audiovisual Archives} (Grant: CRSII5\_198632). The project description is available on:\\\url{https://www.futurecinema.live/project/}

\section*{Impact Statement}

This paper presents work whose goal is to advance the field of machine learning and its applications. There are many potential societal consequences of our work, none which we feel must be specifically highlighted here.

\bibliography{example_paper}
\bibliographystyle{icml2024}

\newpage
\appendix
\onecolumn
\section{Additional Visualizations}

\begin{figure*}[!ht]
    \centering
    \includegraphics[width=\linewidth]{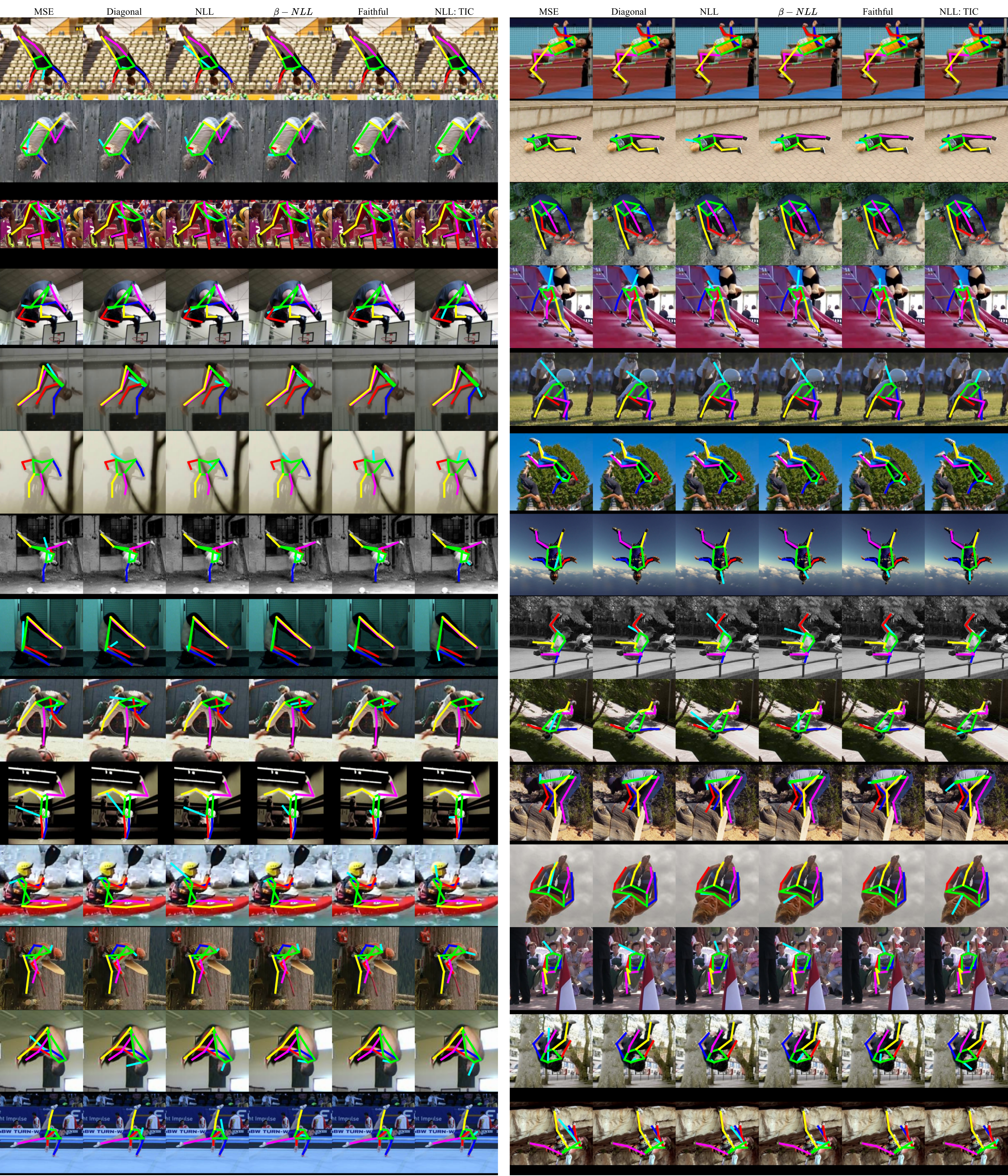}
    \caption{We show additional visualizations to highlight the updated prediction for the head conditioned on observing the ground truth for the remaining joints. TIC accurately updates the location for the head based on successfully learning the correlations underlying the joints.}
    \label{fig:hp_viz_appendix}
\end{figure*}

\section{Human Pose Estimation - Hourglass}

\begin{table*}[h]
  \caption{\textit{Human Pose Results - Stacked Hourglass architecture}. We report the TAC error for each joint along with the average across all joints. Additionally, we report the likelihood score for all methods. We show that NLL-TIC outperforms baselines across all joints and successfully scales to convolutional and transformer based architectures.}
  \label{tab:humanpose_appendix}
  \renewcommand{\arraystretch}{0.9}
  \centering
  \resizebox{\linewidth}{!}{%
  \begin{tabular}{lcccccccccccccccccc}
    \toprule
    Method & head & neck & lsho & lelb & lwri  & rsho & relb & rwri & lhip & lknee &  lankl & rhip & rknee & rankl & \textbf{Avg: TAC} & \textbf{Avg: LL}\\
    \midrule
    MSE &  5.53 & 7.88 &  7.31 &  8.73 & 10.52  & 7.01 & 8.41 &  10.19 &  8.43 & 8.53 & 10.53 & 8.13 & 8.37 & 10.58 & 8.58 $\pm$ 0.21 & -1018.6 $\pm$ 31.2 \\
    NLL-Diagonal & 5.36 &  7.23 &  6.95 &  8.17 &  10.01 & 6.48 & 7.79 &  9.73 & 8.11 & 8.30 &  11.12 &   7.75 & 8.17 &  11.20 & 8.32 $\pm$ 3.19 & -96.3 $\pm$ 127.2 \\
    NLL &  4.48 &  6.81 &  5.38 & 5.19 & 7.13 & 5.11 & 4.86 & 6.89 & 6.62 & 6.35 & 8.45 &  6.43 &  6.17 & 8.40 & 6.31 $\pm$ 0.21 & -93.0 $\pm$ 1.76 \\
    $\beta$-NLL  & 4.63 &   7.14 &  6.74 &   8.23 &   9.98 &  6.43 & 7.92 &  9.65 &   8.01 &  8.13 &  10.12 &  7.71 & 7.93 &  10.19 & 8.06 $\pm$ 0.17 & -97.5 $\pm$ 0.25 \\
    Faithful & 5.13 & 6.36 & 5.32 & 4.94 & 7.18 & 4.96 & 4.72 & 6.85 & 6.67 & 6.29 &  8.39 &  6.36 & 6.22 & 8.37 & 6.27 $\pm 0.06$ & -91.8 $\pm$ 0.22\\
    \midrule
    \textbf{NLL-TIC} &  \textbf{3.76} & \textbf{5.98} & \textbf{4.80} & \textbf{4.64} & \textbf{6.34} & \textbf{4.46} & \textbf{4.41} & \textbf{6.12} &  \textbf{6.09} & \textbf{5.82} & \textbf{7.59} & \textbf{5.79} & \textbf{5.63} & \textbf{7.55} & \textbf{5.64} $\pm$ 0.03 & \textbf{-88.6} $\pm$ 0.08\\
    \bottomrule
  \end{tabular}
  }
\end{table*}



\end{document}

%% file: math_commands.tex
\usepackage{amsmath,amsfonts,bm}

\def\eqref#1{equation~\ref{#1}}

\def\1{\bm{1}}

\def\vmu{{\bm{\mu}}}

\def\vh{{\bm{h}}}

\def\vq{{\bm{q}}}

\def\vx{{\bm{x}}}
\def\vy{{\bm{y}}}
\def\vz{{\bm{z}}}

\def\mI{{\bm{I}}}
\def\mJ{{\bm{J}}}

\def\mSigma{{\bm{\Sigma}}}

\DeclareMathAlphabet{\mathsfit}{\encodingdefault}{\sfdefault}{m}{sl}
\SetMathAlphabet{\mathsfit}{bold}{\encodingdefault}{\sfdefault}{bx}{n}
\newcommand{\tens}[1]{\bm{\mathsfit{#1}}}

\def\tH{{\tens{H}}}

\def\tX{{\tens{X}}}
\def\tY{{\tens{Y}}}

\newcommand{\Var}{\mathrm{Var}}

\newcommand{\Cov}{\mathrm{Cov}}


%% file: example_paper.bbl
\begin{thebibliography}{45}
\providecommand{\natexlab}[1]{#1}
\providecommand{\url}[1]{\texttt{#1}}
\expandafter\ifx\csname urlstyle\endcsname\relax
  \providecommand{\doi}[1]{doi: #1}\else
  \providecommand{\doi}{doi: \begingroup \urlstyle{rm}\Url}\fi

\bibitem[Andriluka et~al.(2014)Andriluka, Pishchulin, Gehler, and Schiele]{mpii}
Andriluka, M., Pishchulin, L., Gehler, P., and Schiele, B.
\newblock 2d human pose estimation: New benchmark and state of the art analysis.
\newblock In \emph{IEEE Conference on Computer Vision and Pattern Recognition (CVPR)}, June 2014.

\bibitem[Bertoni et~al.(2019)Bertoni, Kreiss, and Alahi]{bertoni2019monoloco}
Bertoni, L., Kreiss, S., and Alahi, A.
\newblock Monoloco: Monocular 3d pedestrian localization and uncertainty estimation.
\newblock In \emph{Proceedings of the IEEE/CVF international conference on computer vision}, pp.\  6861--6871, 2019.

\bibitem[Dorta et~al.(2018)Dorta, Vicente, Agapito, Campbell, and Simpson]{dorta2018structured}
Dorta, G., Vicente, S., Agapito, L., Campbell, N.~D., and Simpson, I.
\newblock Structured uncertainty prediction networks.
\newblock In \emph{Proceedings of the IEEE conference on computer vision and pattern recognition}, pp.\  5477--5485, 2018.

\bibitem[Dosovitskiy et~al.(2021)Dosovitskiy, Beyer, Kolesnikov, Weissenborn, Zhai, Unterthiner, Dehghani, Minderer, Heigold, Gelly, Uszkoreit, and Houlsby]{dosovitskiy2021an}
Dosovitskiy, A., Beyer, L., Kolesnikov, A., Weissenborn, D., Zhai, X., Unterthiner, T., Dehghani, M., Minderer, M., Heigold, G., Gelly, S., Uszkoreit, J., and Houlsby, N.
\newblock An image is worth 16x16 words: Transformers for image recognition at scale.
\newblock In \emph{International Conference on Learning Representations}, 2021.
\newblock URL \url{https://openreview.net/forum?id=YicbFdNTTy}.

\bibitem[Dua \& Graff(2017)Dua and Graff]{Dua:2019}
Dua, D. and Graff, C.
\newblock {UCI} machine learning repository, 2017.
\newblock URL \url{http://archive.ics.uci.edu/ml}.

\bibitem[Evans \& Rosenthal(2004)Evans and Rosenthal]{evans2004probability}
Evans, M.~J. and Rosenthal, J.~S.
\newblock \emph{Probability and statistics: The science of uncertainty}.
\newblock Macmillan, 2004.
\newblock URL \url{https://www.utstat.toronto.edu/mikevans/jeffrosenthal/book.pdf}.

\bibitem[Gal \& Ghahramani(2016)Gal and Ghahramani]{gal2016dropout}
Gal, Y. and Ghahramani, Z.
\newblock Dropout as a bayesian approximation: Representing model uncertainty in deep learning.
\newblock In \emph{international conference on machine learning}, pp.\  1050--1059. PMLR, 2016.

\bibitem[Gal et~al.(2017)Gal, Islam, and Ghahramani]{gal2017deep}
Gal, Y., Islam, R., and Ghahramani, Z.
\newblock Deep bayesian active learning with image data.
\newblock \emph{arXiv preprint arXiv:1703.02910}, 2017.

\bibitem[Gilmer et~al.(2022)Gilmer, Ghorbani, Garg, Kudugunta, Neyshabur, Cardoze, Dahl, Nado, and Firat]{gilmer2022a}
Gilmer, J., Ghorbani, B., Garg, A., Kudugunta, S., Neyshabur, B., Cardoze, D., Dahl, G.~E., Nado, Z., and Firat, O.
\newblock A loss curvature perspective on training instabilities of deep learning models.
\newblock In \emph{International Conference on Learning Representations}, 2022.
\newblock URL \url{https://openreview.net/forum?id=OcKMT-36vUs}.

\bibitem[Gou et~al.(2021)Gou, Yu, Maybank, and Tao]{gou2021knowledge}
Gou, J., Yu, B., Maybank, S.~J., and Tao, D.
\newblock Knowledge distillation: A survey.
\newblock \emph{International Journal of Computer Vision}, 129\penalty0 (6):\penalty0 1789--1819, 2021.

\bibitem[Gundavarapu et~al.(2019)Gundavarapu, Srivastava, Mitra, Sharma, and Jain]{ajain}
Gundavarapu, N.~B., Srivastava, D., Mitra, R., Sharma, A., and Jain, A.
\newblock Structured aleatoric uncertainty in human pose estimation.
\newblock In \emph{CVPR Workshops}, volume~2, 2019.

\bibitem[Houlsby et~al.(2011)Houlsby, Husz{\'a}r, Ghahramani, and Lengyel]{houlsby2011bayesian}
Houlsby, N., Husz{\'a}r, F., Ghahramani, Z., and Lengyel, M.
\newblock Bayesian active learning for classification and preference learning.
\newblock \emph{stat}, 1050:\penalty0 24, 2011.

\bibitem[Immer et~al.(2023)Immer, Palumbo, Marx, and Vogt]{immer2023effective}
Immer, A., Palumbo, E., Marx, A., and Vogt, J.~E.
\newblock Effective bayesian heteroscedastic regression with deep neural networks.
\newblock In \emph{Thirty-seventh Conference on Neural Information Processing Systems}, 2023.
\newblock URL \url{https://openreview.net/forum?id=A6EquH0enk}.

\bibitem[Ioffe \& Szegedy(2015)Ioffe and Szegedy]{ioffe}
Ioffe, S. and Szegedy, C.
\newblock Batch normalization: Accelerating deep network training by reducing internal covariate shift.
\newblock pp.\  448--456, 2015.
\newblock URL \url{http://jmlr.org/proceedings/papers/v37/ioffe15.pdf}.

\bibitem[Johnson \& Everingham(2010)Johnson and Everingham]{lsp}
Johnson, S. and Everingham, M.
\newblock Clustered pose and nonlinear appearance models for human pose estimation.
\newblock In \emph{Proceedings of the British Machine Vision Conference}, 2010.
\newblock doi:10.5244/C.24.12.

\bibitem[Johnson \& Everingham(2011)Johnson and Everingham]{lspet}
Johnson, S. and Everingham, M.
\newblock Learning effective human pose estimation from inaccurate annotation.
\newblock In \emph{Proceedings of IEEE Conference on Computer Vision and Pattern Recognition}, 2011.

\bibitem[Kendall \& Gal(2017)Kendall and Gal]{kendall2017uncertainties}
Kendall, A. and Gal, Y.
\newblock What uncertainties do we need in bayesian deep learning for computer vision?
\newblock In \emph{Advances in neural information processing systems}, pp.\  5574--5584, 2017.

\bibitem[Kingma \& Ba(2015)Kingma and Ba]{kingma2014adam}
Kingma, D.~P. and Ba, J.
\newblock Adam: a method for stochastic optimization.
\newblock In \emph{International Conference on Learning Representations}, 2015.

\bibitem[Kreiss et~al.(2019)Kreiss, Bertoni, and Alahi]{kreiss2019pifpaf}
Kreiss, S., Bertoni, L., and Alahi, A.
\newblock Pifpaf: Composite fields for human pose estimation.
\newblock In \emph{Proceedings of the IEEE/CVF conference on computer vision and pattern recognition}, pp.\  11977--11986, 2019.

\bibitem[Kreiss et~al.(2021)Kreiss, Bertoni, and Alahi]{kreiss2021openpifpaf}
Kreiss, S., Bertoni, L., and Alahi, A.
\newblock Openpifpaf: Composite fields for semantic keypoint detection and spatio-temporal association.
\newblock \emph{IEEE Transactions on Intelligent Transportation Systems}, 23\penalty0 (8):\penalty0 13498--13511, 2021.

\bibitem[Lakshminarayanan et~al.(2017)Lakshminarayanan, Pritzel, and Blundell]{lakshminarayanan2017simple}
Lakshminarayanan, B., Pritzel, A., and Blundell, C.
\newblock Simple and scalable predictive uncertainty estimation using deep ensembles.
\newblock \emph{Advances in neural information processing systems}, 30, 2017.

\bibitem[Le et~al.(2005)Le, Smola, and Canu]{gp_heteroscedastic}
Le, Q.~V., Smola, A.~J., and Canu, S.
\newblock Heteroscedastic gaussian process regression.
\newblock In \emph{Proceedings of the 22nd international conference on Machine learning}, pp.\  489--496, 2005.

\bibitem[Li et~al.(2021{\natexlab{a}})Li, Chen, Shi, Lou, Li, and Lu]{li2021localization}
Li, J., Chen, T., Shi, R., Lou, Y., Li, Y.-L., and Lu, C.
\newblock Localization with sampling-argmax.
\newblock \emph{Advances in Neural Information Processing Systems}, 34:\penalty0 27236--27248, 2021{\natexlab{a}}.

\bibitem[Li et~al.(2021{\natexlab{b}})Li, Xu, Chen, Bian, Yang, and Lu]{li2021hybrik}
Li, J., Xu, C., Chen, Z., Bian, S., Yang, L., and Lu, C.
\newblock Hybrik: A hybrid analytical-neural inverse kinematics solution for 3d human pose and shape estimation.
\newblock In \emph{Proceedings of the IEEE/CVF conference on computer vision and pattern recognition}, pp.\  3383--3393, 2021{\natexlab{b}}.

\bibitem[Liu et~al.(2018)Liu, Ok, Vega-Brown, and Roy]{8461047}
Liu, K., Ok, K., Vega-Brown, W., and Roy, N.
\newblock Deep inference for covariance estimation: Learning gaussian noise models for state estimation.
\newblock In \emph{2018 IEEE International Conference on Robotics and Automation (ICRA)}, pp.\  1436--1443, 2018.
\newblock \doi{10.1109/ICRA.2018.8461047}.

\bibitem[Lotfi et~al.(2022)Lotfi, Izmailov, Benton, Goldblum, and Wilson]{lotfi2022bayesian}
Lotfi, S., Izmailov, P., Benton, G., Goldblum, M., and Wilson, A.~G.
\newblock Bayesian model selection, the marginal likelihood, and generalization.
\newblock In \emph{International Conference on Machine Learning}, pp.\  14223--14247. PMLR, 2022.

\bibitem[Lu \& Koniusz(2022)Lu and Koniusz]{lu2022few}
Lu, C. and Koniusz, P.
\newblock Few-shot keypoint detection with uncertainty learning for unseen species.
\newblock In \emph{Proceedings of the IEEE/CVF Conference on Computer Vision and Pattern Recognition}, pp.\  19416--19426, 2022.

\bibitem[Ly et~al.(2017)Ly, Marsman, Verhagen, Grasman, and Wagenmakers]{ly2017tutorial}
Ly, A., Marsman, M., Verhagen, J., Grasman, R.~P., and Wagenmakers, E.-J.
\newblock A tutorial on fisher information.
\newblock \emph{Journal of Mathematical Psychology}, 80:\penalty0 40--55, 2017.

\bibitem[Nakka \& Salzmann(2023)Nakka and Salzmann]{nakka2023understanding}
Nakka, K.~K. and Salzmann, M.
\newblock Understanding pose and appearance disentanglement in 3d human pose estimation.
\newblock \emph{arXiv preprint arXiv:2309.11667}, 2023.

\bibitem[Newell et~al.(2016)Newell, Yang, and Deng]{hg}
Newell, A., Yang, K., and Deng, J.
\newblock Stacked hourglass networks for human pose estimation.
\newblock In Leibe, B., Matas, J., Sebe, N., and Welling, M. (eds.), \emph{Computer Vision -- ECCV 2016}, pp.\  483--499, Cham, 2016. Springer International Publishing.
\newblock ISBN 978-3-319-46484-8.

\bibitem[Nix \& Weigend(1994)Nix and Weigend]{nix1994estimating}
Nix, D.~A. and Weigend, A.~S.
\newblock Estimating the mean and variance of the target probability distribution.
\newblock In \emph{Proceedings of 1994 ieee international conference on neural networks (ICNN'94)}, volume~1, pp.\  55--60. IEEE, 1994.

\bibitem[Petersen \& Pedersen(2012)Petersen and Pedersen]{IMM2012-03274}
Petersen, K.~B. and Pedersen, M.~S.
\newblock The matrix cookbook, nov 2012.
\newblock URL \url{http://www2.compute.dtu.dk/pubdb/pubs/3274-full.html}.
\newblock Version 20121115.

\bibitem[Ronneberger et~al.(2015)Ronneberger, Fischer, and Brox]{unet}
Ronneberger, O., Fischer, P., and Brox, T.
\newblock U-net: Convolutional networks for biomedical image segmentation.
\newblock In \emph{International Conference on Medical image computing and computer-assisted intervention}, pp.\  234--241. Springer, 2015.

\bibitem[Russell \& Reale(2021)Russell and Reale]{russell2021multivariate}
Russell, R.~L. and Reale, C.
\newblock Multivariate uncertainty in deep learning.
\newblock \emph{IEEE Transactions on Neural Networks and Learning Systems}, 33\penalty0 (12):\penalty0 7937--7943, 2021.

\bibitem[Seitzer et~al.(2022)Seitzer, Tavakoli, Antic, and Martius]{seitzer2022on}
Seitzer, M., Tavakoli, A., Antic, D., and Martius, G.
\newblock On the pitfalls of heteroscedastic uncertainty estimation with probabilistic neural networks.
\newblock In \emph{International Conference on Learning Representations}, 2022.
\newblock URL \url{https://openreview.net/forum?id=aPOpXlnV1T}.

\bibitem[Shukla(2022)]{shukla2022bayesian}
Shukla, M.
\newblock Bayesian uncertainty and expected gradient length-regression: Two sides of the same coin?
\newblock In \emph{Proceedings of the IEEE/CVF Winter Conference on Applications of Computer Vision}, pp.\  2367--2376, 2022.

\bibitem[Shukla et~al.(2022)Shukla, Roy, Singh, Ahmed, and Alahi]{shukla2022vl4pose}
Shukla, M., Roy, R., Singh, P., Ahmed, S., and Alahi, A.
\newblock Vl4pose: Active learning through out-of-distribution detection for pose estimation.
\newblock In \emph{Proceedings of the 33rd British Machine Vision Conference}, number CONF. BMVA Press, 2022.

\bibitem[Shynk(2012)]{shynk2012probability}
Shynk, J.~J.
\newblock \emph{Probability, random variables, and random processes: theory and signal processing applications}.
\newblock John Wiley \& Sons, 2012.

\bibitem[Simpson et~al.(2022)Simpson, Vicente, and Campbell]{simpson2022learning}
Simpson, I.~J., Vicente, S., and Campbell, N.~D.
\newblock Learning structured gaussians to approximate deep ensembles.
\newblock In \emph{Proceedings of the IEEE/CVF Conference on Computer Vision and Pattern Recognition}, pp.\  366--374, 2022.

\bibitem[Skafte et~al.(2019)Skafte, J{\o}rgensen, and Hauberg]{skafte2019reliable}
Skafte, N., J{\o}rgensen, M., and Hauberg, S.
\newblock Reliable training and estimation of variance networks.
\newblock \emph{Advances in Neural Information Processing Systems}, 32, 2019.

\bibitem[Stirn et~al.(2023)Stirn, Wessels, Schertzer, Pereira, Sanjana, and Knowles]{stirn2023faithful}
Stirn, A., Wessels, H., Schertzer, M., Pereira, L., Sanjana, N., and Knowles, D.
\newblock Faithful heteroscedastic regression with neural networks.
\newblock In \emph{International Conference on Artificial Intelligence and Statistics}, pp.\  5593--5613. PMLR, 2023.

\bibitem[Sun et~al.(2019)Sun, Xiao, Liu, and Wang]{hrnet}
Sun, K., Xiao, B., Liu, D., and Wang, J.
\newblock Deep high-resolution representation learning for human pose estimation.
\newblock In \emph{Proceedings of the IEEE/CVF Conference on Computer Vision and Pattern Recognition (CVPR)}, June 2019.

\bibitem[Tekin et~al.(2017)Tekin, Marquez-Neila, Salzmann, and Fua]{Tekin_2017_ICCV}
Tekin, B., Marquez-Neila, P., Salzmann, M., and Fua, P.
\newblock Learning to fuse 2d and 3d image cues for monocular body pose estimation.
\newblock In \emph{Proceedings of the IEEE International Conference on Computer Vision (ICCV)}, Oct 2017.

\bibitem[Xu et~al.(2022)Xu, Zhang, Zhang, and Tao]{xu2022vitpose}
Xu, Y., Zhang, J., Zhang, Q., and Tao, D.
\newblock Vitpose: Simple vision transformer baselines for human pose estimation.
\newblock \emph{Advances in Neural Information Processing Systems}, 35:\penalty0 38571--38584, 2022.

\bibitem[Yao et~al.(2020)Yao, Gholami, Keutzer, and Mahoney]{yao2020pyhessian}
Yao, Z., Gholami, A., Keutzer, K., and Mahoney, M.~W.
\newblock Pyhessian: Neural networks through the lens of the hessian.
\newblock In \emph{2020 IEEE international conference on big data (Big data)}, pp.\  581--590. IEEE, 2020.

\end{thebibliography}
